# Stanford Doggo: An Open-Source, Quasi-Direct-Drive Quadruped

Nathan Kau, Aaron Schultz, Natalie Ferrante, Patrick Slade

*Abstract*— This paper presents Stanford Doggo, a quasi-direct-drive quadruped capable of dynamic locomotion. This robot matches or exceeds common performance metrics of state-of-the-art legged robots. In terms of vertical jumping agility, a measure of average vertical speed, Stanford Doggo matches the best performing animal and surpasses the previous best robot by 22%. An overall design architecture is presented with focus on our quasi-direct-drive design methodology. The hardware and software to replicate this robot is open-source, requires only hand tools for manufacturing and assembly, and costs less than $3000.

## I. INTRODUCTION

Legged robots provide a highly mobile platform to traverse difficult terrain and are ideal for accomplishing tasks that are repetitive, strenuous, or dangerous. Many state-of-the-art legged robots have achieved remarkable feats including high speed locomotion [1], [2], agile maneuvers [3], [4], [5], and traversing difficult terrain [6], [7], [8], [9]. Some designs store energy, such as with a parallel-elastic leg mechanism, to achieve dynamic motion and are not capable of continuously agile motion required to accomplish many tasks [10], [11], [12]. Only robots that carry their power supply and are capable of repeated jumping or locomotion are considered in this work. These platforms take years of development and are frequently expensive, custom designs.

Many metrics characterize the performance of legged robots: steady velocity during running, jump height, and vertical jumping agility. Vertical jumping agility quantifies how quickly an animal can change its energetic state, approximating the vertical climbing speed through a series of jumps [5], [13]. This metric correlates to locomotion performance as the distance a robot can jump increases its ability to overcome obstacles, improving path-planning capabilities [14]. While current quadruped robots are a popular platform capable of carrying payloads, performing manipulation, and fall recovery, they are unable to match the vertical jumping agility of specialized monopod robots [5], [15] that attempt to emulate the animal with the best jumping agility, the galago (*Galago senegalensis*) [16].

Legged robots require a trade-off between energy efficiency to accomplish their task and sensitivity to safely interact with their environment. Often, robots employ a high reduction gear train to increase the effective torque produced by the motor. This rigid gear train requires compliance to be designed in series with the motor, referred to as a series-elastic actuator [17]. Direct-drive (DD) robots do not employ any speed reduction between the motor and output

The authors are with the Department of Mechanical Engineering, Stanford University, Stanford, CA 94305 USA (e-mail: nathankau@stanford.edu).

shaft, allowing the sensitive motor to implement tunable compliance through control at kHz timescales and maintain a larger control bandwidth than the series-elastic actuators [18]. A compromise is achieved with a quasi-direct-drive (QDD) which uses a single stage reduction with a ratio less than 10:1. This single stage increases torque output at the expense of control bandwidth, but maintains the ability to backdrive the motor which allows sensing of external forces based on motor current [19].

This paper presents Stanford Doggo (Fig. 1), a QDD quadruped that matches or exceeds common performance metrics of state-of-the-art legged robots. The mechanical design utilizes a belt drive as a lightweight QDD transmission to increase the effective torque while maintaining low reflected inertia and transparency to enable sensitive control. The QDD enables Stanford Doggo to match the performance of the animal with the best vertical specific agility [16], a 22% improvement over the previous best robot [15]. Stanford Doggo uses completely open-source hardware and software, enabling full replication [20]. The total cost of materials and machining costs is less than $3000 and requires only hand tools for assembly. We hope to advance research and education in the field of legged robotics by lowering the cost and resources required to have access to a state-of-the-art robot.

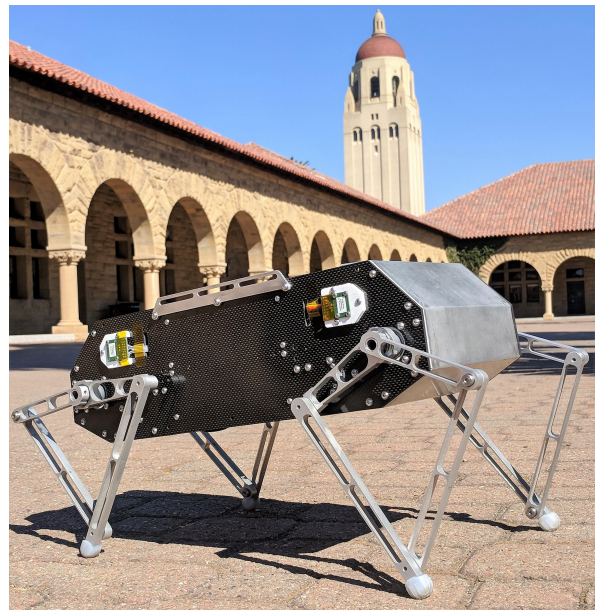

Fig. 1: Stanford Doggo: an open-source, quasi-direct-drive quadruped robot.

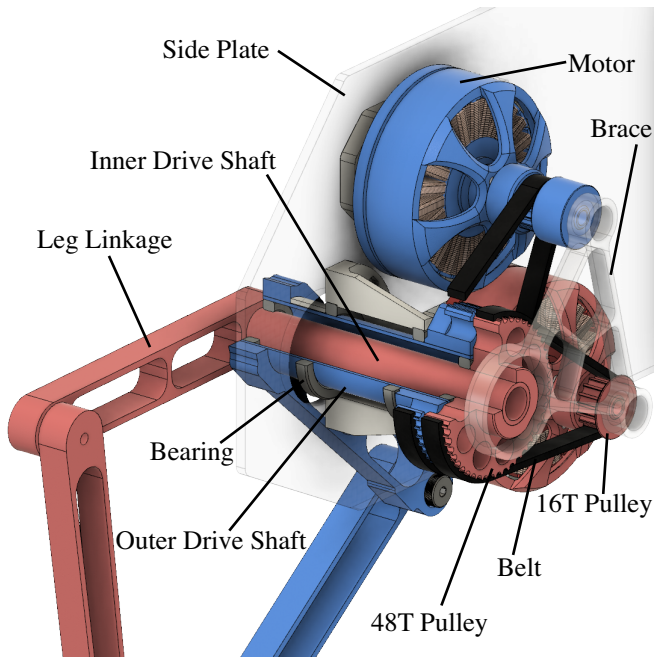

Fig. 2: Assembly of the mechanical components in one leg. Red and blue colors indicate parts corresponding to one belt drive from motor to leg linkage.

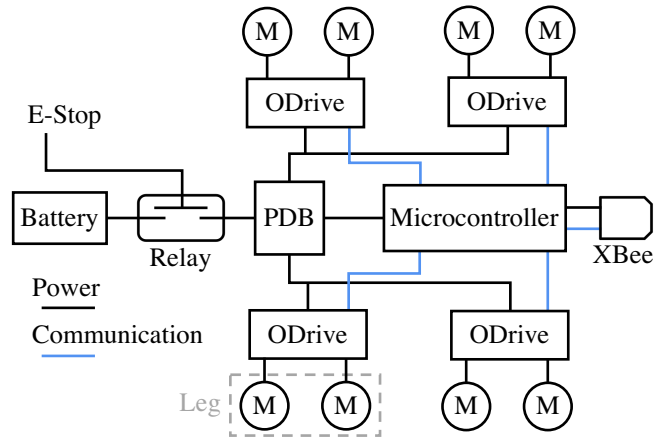

Fig. 3: Electrical diagram for motor control and power system.

## II. METHODS

### A. Design

In order to allow others to replicate Stanford Doggo we made the design open-source [20]. The project repository includes all CAD files, a detailed bill of materials, general assembly instructions, wiring schematics, and all necessary software. The total cost estimated to build Stanford Doggo is $3000. This includes estimated costs to have outsourced machining services perform any manufacturing beyond those possible with hand tools, such as a hand drill or soldering iron. In order to increase accessibility, all components and quoted machining services are available online. The overall dimensions of the Stanford Doggo chassis are 42cm in length, 20cm in width, and 14cm in height. The minimum and maximum leg extensions, measured from the foot to the center of rotation are 8cm and 25cm, respectively.

Stanford Doggo features a QDD transmission that improves upon the mechanical design of Minitaur, another quadruped with a similar leg linkage design [4]. Minitaur mounts the two motors used to control each leg on either side of the leg linkage. However, the structural element connecting these two motors prevents the leg from completing a full revolution and limits the workspace of the robot. In contrast, the Stanford Doggo coaxial drive assemblies allows the legs to rotate outside of the body without any constraints (Fig. 2). Each coaxial drive assembly contains two 3:1 belt drives which transfer power from the motors to the drive shafts. The two drive shafts nest inside each other coaxially which saves both space and weight while allowing the leg to rotate continuously in either direction.

A belt drive was selected as the QDD mechanism over a geared system to minimize weight, cost, and reflected inertia. While the MIT Cheetah 2 and other quadrupeds used planetary gears systems to increase torque, these systems usually require heavy steel gears to support the loads involved when high torque is applied [21]. In contrast, the belt drive on Stanford Doggo distribute loads across several teeth, which reduces the tooth load and enables the use of lightweight, plastic materials. The use of plastic also reduces the inertia of the pulleys which improves transparency. The pulleys are easily fabricated on a 3D printer which lowers the cost of manufacturing.

The electrical system of Stanford Doggo consists of a microcontroller, four leg subsystems containing two motors (M) and a motor controller (ODrive), a power distribution board (PDB), and a communications system (Fig. 3). A wireless module is used to send commands from a groundstation to the Teensy 3.5 microcontroller (PJRC, Sherwood, OR). The Teensy 3.5 was selected because it is a low-cost, Arduino-compatible microcontroller that is common among the open-source community. The microcontroller computes leg trajectories and sends leg position commands to the ODrive motor controllers (ODrive Robotics, Richmond, CA) at 100Hz. The motor controllers run field-oriented-control motor commutation at 10kHz to control the torque applied by the MN5212 motors (Tiger Motor, Nanchang, China) with positional feedback from axially mounted magnetic encoders with 2000 counts per revolution. A relay is connected in series between the batteries and the PDB with a push button to serve as an emergency stop to disconnect power to the robot.

Stanford Doggo achieves walking, trotting, bounding, and pronking gaits by commanding sinusoidal open-loop trajectories to the four motor controllers. Minitaur was able to generate stable gaits using a similar open-loop trajectory method [22]. The leg trajectories used on Stanford Doggo are composed of two halves of sinusoidal curves for the flight and stance phases shown in orange and purple in Fig. 5.

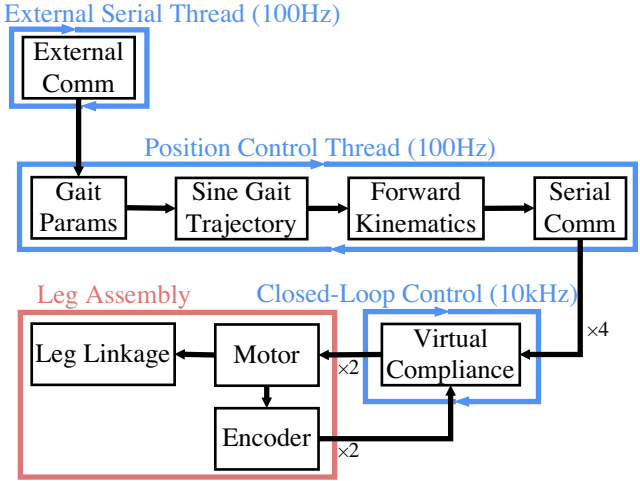

Fig. 4: Control framework for the Stanford Doggo.

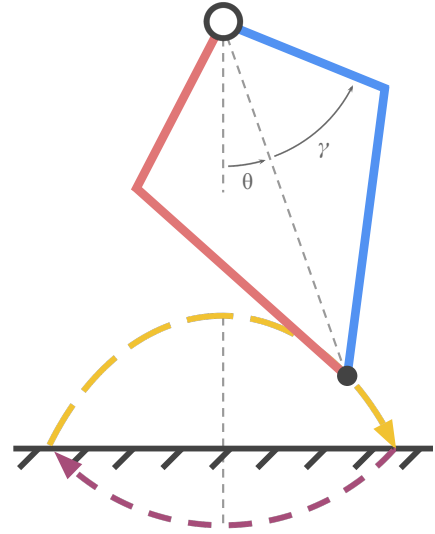

Fig. 5: Piecewise sinusoids were used to design different gaits by varying the relative amplitudes, frequencies, stride length, and virtual compliance of the leg.

The geometric parameters of the sinusoids, the virtual leg compliance, and the duration of time that the leg spends traversing each sinusoidal segment, were varied to create different gaits. These trajectories are converted to a desired leg angle ($\theta$) and a desired link separation ($\gamma$) that describe the virtual leg that originates at the hip joint of the leg and terminates at the foot. These virtual leg parameters and their corresponding stiffness and damping coefficients are sent from the microcontroller to the motor controllers. The motor controllers use PD control to generate output torques to achieve the desired leg angle and separation. These generalized torques are converted to motor torques by using the relationship with the kinematic leg Jacobian. The proportional term of the controller adds virtual stiffness and the derivative term adds virtual damping. Taken together, the two terms create a virtually compliant system.

### B. Transmission Principles

The design of Stanford Doggo's QDD transmission was informed by the scaling laws for actuator torque, transparency, and weight. For a QDD transmission to be advantageous over a direct-drive design, the QDD device should increase the output torque while keeping the additional mass small enough such that a larger motor would not provide the same benefit. Similarly, the reflected inertia introduced by the QDD transmission should be kept low enough to maintain sensitive proprioception.

Taking these criteria into account, a threshold mass can be calculated for the QDD transmission to be beneficial over a direct-drive design. If an electromagnet motor's radius ($r$) is scaled, motor mass is $\propto r$ assuming that the mass is concentrated near the air-gap of the motor. Additionally, torque is $\propto r^2$, and reflected inertia is $\propto r^3$ [23]. For a QDD transmission with a speed reduction of $N$:1, the torque multiplier is $\propto N$. The output torque can either be increased by choosing a larger motor or by using a higher reduction. A larger motor that increases the gap radius by a factor of $\sqrt{N}$ would achieve approximately the same output torque as a QDD transmission with a speed reduction of ratio $N$:1, while also multiplying the motor mass by a factor of $\sqrt{N}$. The QDD transmission mass should thus be less than $(\sqrt{N}-1)m$, where $m$ is the motor mass, in order to save mass compared to a larger motor that produces the same torque. In practice, this mass criteria can be met by various QDD designs, but may reduce the control bandwidth, efficiency, and reliability of the system. In terms of transparency, which is important to sensitive force production and compliance, reflected inertia is invariant to the choice of QDD or DD [4]. However, transparency is still negatively affected as $N$ increases due to dynamics in the transmission. These trade-offs should be considered for the specific application.

### C. Performance Metrics

Stanford Doggo's transmission and motor were characterized with a number of metrics to understand the trade-offs between cost, torque, mass of the motor plus transmission, power, and reflected inertia. The maximum continuous torque was measured by attaching a lever arm to the motor, placing a scale under the opposite end of the lever arm, and commanding the motor to exert force against the scale. Motor current was then gradually increased until the motor reached a steady state temperature of 125°C, a value often used as the maximum permissible temperature of stator windings [24]. During the experiment the ambient temperature was 22°C and no cooling was applied to the motor. Once the maximum continuous current was determined, the force against the scale was multiplied by the link length and by the transmission speed reduction ratio to find the maximum continuous torque at the output. The peak torque was measured using the same experimental setup and was based on the peak force exerted over a one second duration. The reflected inertia of the motor was estimated numerically by using a computer-aided-design model of the motor's rotor. This inertia value was then multiplied by the square of the speed reduction ratio to find the reflected inertia. The torque density, which is the ratio of peak motor torque to motor mass, was analyzed to understand the robot's ability to exert force, one of the limiting resources on a legged robot.

The transparency of the leg assembly was investigated by comparing the predicted foot force to the measured foot force. The leg was constrained such that the foot was located under the hip and connected in tension with a LC201 load cell (Omega Engineering, Norwalk, CT). An analog to digital converter provided readings at 10Hz from the load cell. Only the steady state error was considered due to the slow sampling speed. The force ($F_{est}$) was estimated by using the leg Jacobian ($J$) to convert applied motor torques to a force vector acting on the foot. The Jacobian for this parallel linkage configuration has been detailed thoroughly [22]. The motor torques could not be directly measured so they were estimated by using the motor torque constant ($K_t$) which describes the linear relationship between motor currents ($I_m$) and motor torques ($\tau_m$),

$$\tau_m = K_t I_m \quad (1)$$

$$F_{est} = J^{-T} \tau_m. \quad (2)$$

The virtual compliance of the leg was controlled to increase and decrease the vertical foot force while maintaining the load cell in tension to prevent loading and unloading motions.

The control bandwidth was estimated using the same experimental setup. Sinusoidal torque trajectories with a minimum and maximum torque of 4Nm and 12Nm were tracked across a range of frequencies from 5Hz to 400Hz. This torque range was selected to emulate the torques seen during walking. The analog to digital converter provided readings at 18kHz. The amplitude and phase of the response during steady state tracking were recorded. These values were used to find the crossover frequency which approximates the bandwidth.

General physical properties of state-of-the-art legged robots are compared to give intuition to important characteristics in the design of legged robots. These properties included number of legs, total degrees of freedom for the robot (DOF), leg length, mass of the robot, percentage of the total robot mass contributed by motors without the weight of the transmission included, and the speed reduction ratio.

To help understand the trade-offs between agility and efficiency, common performance metrics were computed using previously defined methods [4], [5]. Steady velocity ($v_{ss}$) defines the maximum velocity recorded during forward running, representing the ability to quickly traverse flat terrain. This velocity was computed as the time taken to run 0.7 meters over ground. The running trial was started once the the steady velocity was reached.

TABLE I: Comparison of QDD and DD Actuators

| Robot | Stanford Doggo | Minitaur [4] |
| --- | --- | --- |
| Actuator | T-Motor MN5212, 3:1 | T-Motor U8 |
| Cost | $120 | $280 |
| Speed Reduction | 3:1 | 1:1 |
| Mass (kg) | 0.27 | 0.25 |
| Continuous Torque (Nm) | 1.51 | 0.86 |
| Peak Torque (Nm) | 4.8 | 3.5 |
| Max Continuous Power (W) | 840 | 179 |
| Reflected Inertia (kg·m$^2$) | 0.00026 | 0.0001 |

The cost of transport is presented to inform the efficiency of this fast forward locomotion. The cost of transport is computed from the mean voltage ($V$) and current ($i$) applied during steady velocity running as well as the mass ($m$) of the robot,

$$\text{Cost of Transport} := \frac{Vi}{mgv_{ss}}. \quad (3)$$

The peak vertical jump height is measured as the maximum vertical height achieved in a jump measured from the robot's center-of-gravity in the crouched position to the center-of-gravity at the apex of the jump. The authors of [5], [15] were contacted to confirmed this method of measurement because prior work that reported these metrics did not specify.

Vertical jumping agility quantifies how quickly an animal can change its energetic state and approximates the speed at which a system could climb vertically through a series of jumps [13], [5]. Vertical jumping agility is computed as the peak vertical jump height divided by the combined time in stance and time to the apogee. The time in stance is defined as the time taken from the initiation of actuation starting the jump to the instant the feet leave the ground. The time to the apogee is the recorded from when the feet leave the ground until the apogee of the jump, where the vertical velocity is zero.

### III. RESULTS

Careful consideration was given to the actuator and transmission selection which determined the trade-off between torque, transparency, and mass. A comparison with the Minitaur's actuator (Table I) shows that the Stanford Doggo's actuator has a 37% higher peak force and a 76% higher continuous torque at the cost of an 8% increase in total mass and three times increase in inertia. The total cost of the actuator is also significantly less than the cost of the Minitaur actuator. The MIT Cheetah 2 uses a QDD transmission with a mass roughly equal to the motor and a torque density of $58\frac{\text{Nm}}{\text{kg}}$ [25]. In comparison, the mass of the Stanford Doggo belt drive assembly is 27% of the motor mass and the mechanism has a torque density of $17.8\frac{\text{Nm}}{\text{kg}}$.

The transparency of the QDD transmission was investigated by tracking force trajectories for step functions and 2 rad/s sine waves. These force trajectories had mean absolute percent errors of 2.8% and 5.3%, respectively (Fig. 6). The control bandwidth was estimated to be 150Hz for torque trajectories that emulated walking.

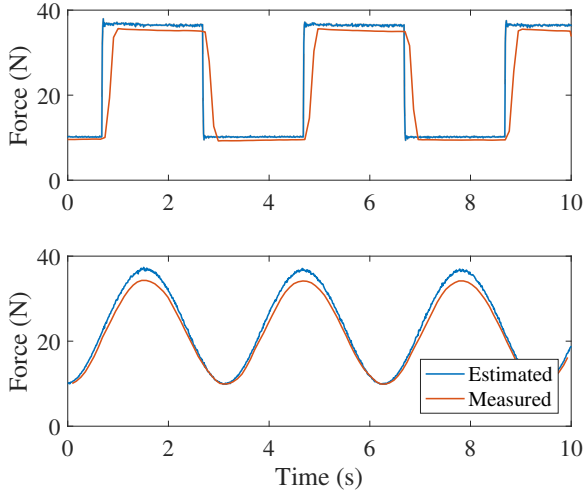

Fig. 6: A comparison of the force applied by the leg as measured by a load cell and the force estimated from the kinematics and motor currents.

TABLE II: Physical Properties of Legged Robots

| Robot | Legs | DOF | Leg Length (m) | Mass (kg) | Mass From Motors (%) | Gear Ratio |
| --- | --- | --- | --- | --- | --- | --- |
| Stanford Doggo | 4 | 8 | 0.160 | 4.8 | 35 | 3 |
| Minitaur [4] | 4 | 8 | 0.200 | 5.0 | 40 | 1 |
| Salto-1P [15] | 1 | 4 | 0.144 | 0.1 | 12 | 25 |
| Jerboa [4], [5] | 2 | 4 | 0.105 | 2.5 | 40 | 1 |
| MIT Cheetah 2 [25], [4] | 4 | 12 | 0.275 | 33.0 | 24 | 5.8 |
| MIT Cheetah 3 [26] | 4 | 12 | 0.34 | 45.0 | N/A | 7.67 |
| StarlETH [7], [4] | 4 | 12 | 0.200 | 23.0 | 16 | 100 |
| ANYMAL [8] | 4 | 12 | 0.250 | 30.0 | N/A | N/A |
| Cheetah Cub [27], [4] | 4 | 8 | 0.069 | 1.0 | 16 | 300 |
| XRL [4], [5] | 6 | 6 | 0.200 | 23.0 | 16 | 23 |

The degrees of freedom and leg lengths closely match among many of the compared robots (Table II). There was a wide range of total masses, percentage of the mass contributed by motors, and speed reduction ratios depending on the type of transmission. The motor percentage of Stanford Doggo (35%) is comparable to the motor mass percentage of the other DD robots, Minitaur and Jerboa. However, compared to MIT Cheetah 2, which also uses a QDD leg mechanism, the motor mass percentage is 46% higher, indicating that a significantly larger proportion of Stanford Doggo's mass is committed to torque production rather than structure or electronics.

Stanford Doggo performs adequately in terms of steady velocity and cost of transport in comparison to the other robots (Table III). Salto-1P had a cost of transport roughly twice as large as Stanford Doggo at a four times higher steady velocity. Stanford Doggo achieved the highest jump of any other quadruped by more than two times. The Salto-1P's jump was only 0.11m more than Stanford Doggo, despite being a specialized monopod. Stanford Doggo reported the highest vertical jumping agility among all robots and matched the performance of the galago, the animal with highest known vertical jumping agility [16]. MIT Cheetah

TABLE III: Performance Measures of Legged Robots

| Robot | Steady Velocity (m/s) | Cost of Transport | Maximum Jump Height (m) | Vertical Jumping Agility (m/s) |
| --- | --- | --- | --- | --- |
| Stanford Doggo | 0.9 | 3.2 | 1.14 | 2.23 |
| Minitaur [4] | 1.5 | 2.3 | 0.48 | 1.12 |
| Salto-1P [15] | 3.6 | 6.6 | 1.25 | 1.83 |
| Jerboa [4], [5] | 1.52 | 2.5 | 0.18 | 0.81 |
| MIT Cheetah 2 [25], [4] | 6.0 | 0.51 | 0.50 | 1.11 |
| MIT Cheetah 3 [26] | N/A | 0.45 | N/A | N/A |
| StarlETH [7], [4] | 0.7 | 2.57 | 0.01 | 0.72 |
| ANYMAL [8] | 0.8 | 1.23 | N/A | N/A |
| Cheetah Cub [27], [4] | 1.4 | 9.8 | 0.02 | N/A |
| XRL [4], [5] | 1.54 | 2.57 | 0.31 | 0.92 |

3 produces significantly larger force per weight than MIT Cheetah 2, indicating potential for fast steady velocity and large maximum jump height, but few performance details have been published [26].

## IV. DISCUSSION

The Stanford Doggo actuator uses a QDD mechanism to increase the output torque by a factor of 3. Following the actuator scaling laws, a DD actuator with same output torque would be 73% heavier than the motor in the QDD mechanism. The approximate mass reduction using the QDD totals 770g, or 16% of the total robot mass. Compared to the Minitaur actuator, the Stanford Doggo actuator has 2.7 times the reflected inertia of the Minitaur actuator, yet provides 1.8 times more continuous torque with similar mass [4]. Again following the scaling laws for increasing the gap radius of a DD actuator, attaining a 1.8 times greater torque would require approximately a 1.3 times larger gap radius, which would increase the motor inertia by a factor of 2.4. However, a 1.3 times larger gap radius would increase the motor mass by the same factor of 1.3, which would make the DD actuator 20% heavier than the Stanford Doggo's QDD actuator.

The transparency of the QDD transmission displays accurate force sensing for a variety of commanded trajectories in Fig. 6. Accurate force estimation is essential when interacting with the environment and stabilizing the robot. The control bandwidth was approximated to be 150Hz, higher than the MIT Cheetah 2 which reports a bandwidth between the motor and foot force as 104Hz [28]. The increased bandwidth offers improved control for dynamic maneuvering.

Stanford Doggo's steady velocity was limited by the instability of the open-loop tracking used to control foot position trajectories (Table III). A trotting gait was used for stability at high speeds. Most of the quadrupeds reported use a bounding gait to achieve the highest speed [4]. Since the motor performance was not a limiting factor, a significantly higher peak velocity is expected to be possible with closed-loop control methods [29].

The cost of transport of Stanford Doggo is higher than the Minitaur, which is to be expected due to the inefficient open-loop trajectories used for control. Stanford Doggo will likely see significant improvements with better control [29].

The maximum vertical jump height of Stanford Doggo can be accurately evaluated using open-loop control, and highlights the benefit of the increased torque and lightweight belt drive by more than doubling the jump height of the quadruped robots. This massive performance gap suggests that with sufficient control, Stanford Doggo will be able to outperform most or all quadrupeds in challenging legged locomotion. The reliability of the system proved to be quite robust, undergoing hours of running tests and approximately 40 jumps while requiring only minimal maintenance to adjust belt tension.

Stanford Doggo's vertical jumping agility outperforms all legged robots and highlights the benefits of the increased peak torque in comparison to the Minitaur actuator (Table III). While the MIT Cheetah 2 reported a higher torque density than Stanford Doggo, the increase did not result in a higher vertical jumping agility. Three factors contribute to Stanford Doggo's high vertical jumping agility compared to the MIT Cheetah 2. First, the smaller leg length on Stanford Doggo means that the joint torque acts over a shorter lever arm, which increases the force output at the end of the linkage. Second, the lightweight belt drive transmission in comparison to MIT Cheetah 2's planetary gears system significantly reduces the total mass of the robot [2]. Finally, Stanford Doggo uses a parallel linkage, which, with joint torques and size kept equal, produces a greater maximum foot force than a similarly sized serial linkage because the torques sum together. The high vertical jumping agility of Stanford Doggo shows the potential of Stanford Doggo to perform extremely dynamic maneuvers.

## V. Conclusion

In this work we introduce Stanford Doggo, a robot that merges the dexterity and inherent stability of quadruped robots with a vertical jumping agility greater than specialized monopods and matching that of the highest performing animal, the galago. Stanford Doggo also meets or exceeds state-of-the-art legged robotic systems in a number of common performance metrics. The complete design is open-source with a focus on low manufacturing cost, less than $3000, and requires only hand tools for assembly [20]. In making an accessible, state-of-the-art legged robot platform, we hope to improve research and education in legged robotics by lowering the barriers to entry. We plan to continue the open-source development to improve the design and closed-loop gait controllers. In the future, we intend to explore additional control schemes to fully utilize Stanford Doggo's extreme mobility.

## VI. Acknowledgements

The authors would like to thank Ademi Adeniji, Brian Zeng, James Wang, and Meera Radhakrishnan for supporting the development of Stanford Doggo.